\documentclass[a4paper,twoside]{article}

\usepackage{epsfig}
\usepackage{subcaption}
\usepackage{calc}
\usepackage{amssymb}
\usepackage{amstext}
\usepackage{amsmath}
\usepackage{amsthm}
\usepackage{multicol}
\usepackage{pslatex}
\usepackage{apalike}
\usepackage{algorithm2e}
\usepackage[bottom]{footmisc}
\usepackage{SCITEPRESS}     

\begin{document}

\title{SLIM-RAFT: A Novel Fine-Tuning Approach to Improve Cross-Linguistic Performance for Mercosur Common Nomenclature}

\author{\authorname{Vinícius Di Oliveira\sup{1,2}\orcidAuthor{0000-0002-1295-5221}, Yuri Façanha Bezerra\sup{1}\orcidAuthor{0000-0000-0000-0000}, Li Weigang\sup{1}\orcidAuthor{0000-0003-1826-1850} Pedro Carvalho Brom\sup{1,3}\orcidAuthor{0000-0002-1288-7695} and Victor Rafael R. Celestino\sup{4}\orcidAuthor{0000-0002-1288-7695}}
\affiliation{\sup{1}TransLab, University of Brasilia, Brasilia, Federal District, Brazil}
\affiliation{\sup{2}Secretary of Economy, Brasilia, Federal District, Brazil}
\affiliation{\sup{3}Instituto Federal de Brasilia, Brasilia, Federal District, Brazil}
\affiliation{\sup{4}LAMFO, Department of Administration, University of Brasilia, Brasilia, Federal District, Brazil}
\email{vinidiol@gmail.com, weigang@unb.br}
}


\abstract{Natural language processing (NLP) has seen significant advancements with the advent of large language models (LLMs). However, substantial improvements are still needed for languages other than English, especially for specific domains like the applications of Mercosur Common Nomenclature (NCM), a Brazilian Harmonized System (HS). To address this gap, this study uses TeenyTineLLaMA, a foundational Portuguese LLM, as an LLM source to implement the NCM application processing. Additionally, a simplified Retrieval-Augmented Fine-Tuning (RAFT) technique, termed SLIM-RAFT, is proposed for task-specific fine-tuning of LLMs. This approach retains the chain-of-thought (CoT) methodology for prompt development in a more concise and streamlined manner, utilizing brief and focused documents for training. The proposed model demonstrates an efficient and cost-effective alternative for fine-tuning smaller LLMs, significantly outperforming TeenyTineLLaMA and ChatGPT-4 in the same task. Although the research focuses on NCM applications, the methodology can be easily adapted for HS applications worldwide.
}

\onecolumn \maketitle \normalsize \setcounter{footnote}{0} \vfill

\section{\uppercase{Introduction}}
\label{sec:introduction}

The widespread application of Generative Artificial Intelligence (GenAI) systems has significantly advanced the development of artificial intelligence \cite{schulhoff2024prompt,radosavovic2024humanoid}. On one hand, more sophisticated large language models (LLMs), such as ChatGPT, possess multilingual capabilities to process multimodal information and have become key drivers for AI applications \cite{weigang2022new}. However, for most users, effective use of these models requires enhancing prompt engineering skills. On the other hand, open-source large language models, such as LLaMA 3, can locally fine-tune models, catering to specific security and flexibility needs. Nonetheless, non-English users may encounter limitations due to the composition of LLaMA's training corpus, predominantly English-based (90\%), with only a small portion (10\%) dedicated to other languages, including French, Spanish, and Portuguese.


While LLMs offer cross-language processing capabilities, enabling some understanding of related languages, particularly within the Latin language group, these capabilities are often insufficient for more nuanced language processing tasks \cite{souza2020bertimbau}. This is especially true for tasks involving technical terminology and enterprise-specific privacy data requirements. The limited proportion of pre-trained Portuguese corpus in models like LLaMA highlights these constraints, revealing significant limitations in their language processing capabilities for non-English languages.

\begin{figure*}[!h]
  \centering
   {\epsfig{file = 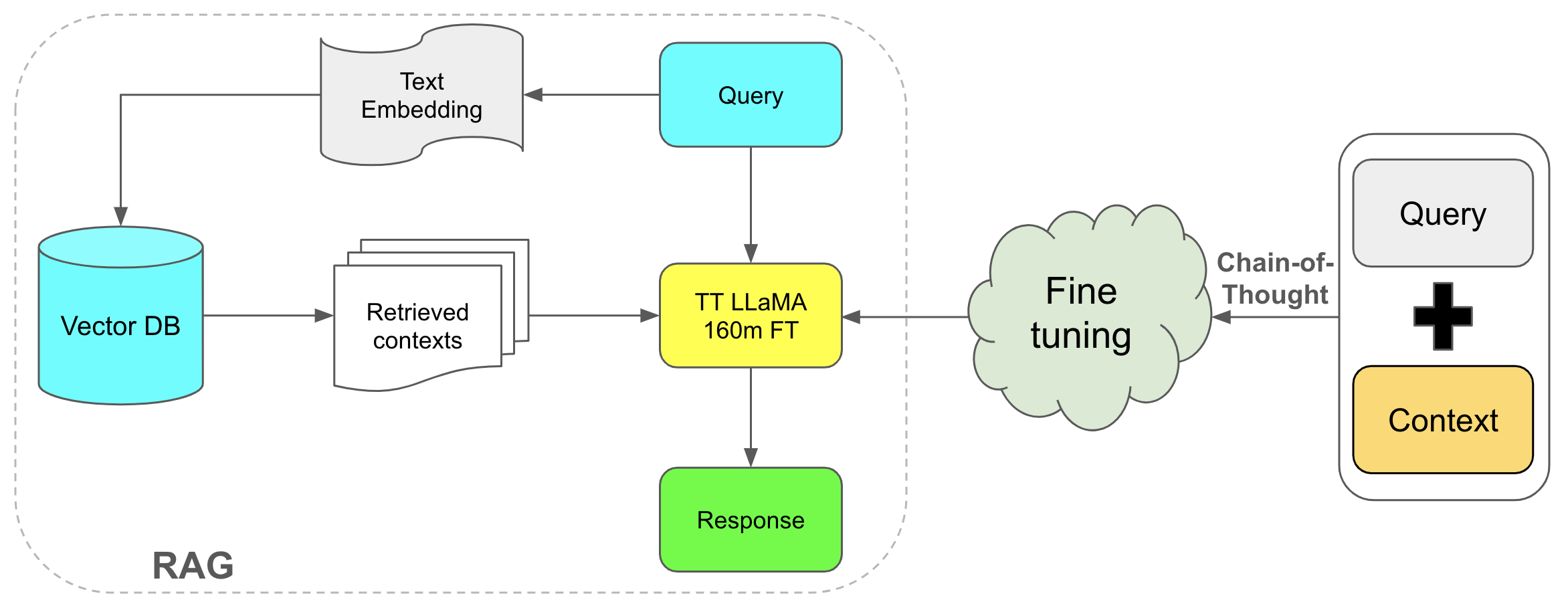, width = 16cm}}
  \caption{The SLIM-RAFT diagram}
  \label{fig: fig1_slim-raft}
 \end{figure*}

Another way to tackle these downstream tasks could be to use significantly smaller LLMs and perform fine-tuning by retraining the model parameters, as demonstrated in the TeenyTineLLaMA model \cite{correa2024teenytinyllama}, which is the approach addressed in this work. This approach considers that subsequent tasks have a well-defined and reduced scope but preserves the essence of using LLMs.

Retrieval-Augmented Generation (RAG) addresses several critical challenges inherent in LLMs \cite{lewis2020rag,gao2023retrieval}. These challenges encompass hallucinations, outdated knowledge, and non-transparent, untraceable reasoning processes. By incorporating information from external databases, RAG significantly enhances the accuracy and credibility of generated content, particularly for knowledge-intensive tasks. This approach facilitates continuous updates of knowledge and the integration of domain-specific information, effectively merging the intrinsic capabilities of LLMs with the extensive, dynamic repositories of external databases. Consequently, RAG provides a robust solution for improving the performance and reliability of language models, making it an invaluable approach across a diverse array of applications.

In the pursuit of enhancing the performance of LLMs in downstream applications, Retrieval-Augmented Fine-Tuning (RAFT) presents numerous significant benefits \cite{zhang2024raft,warnakulasuriyaknowledge}. While pre-training LLMs on extensive corpora of textual data are now standard practice, integrating new, time-sensitive, or domain-specific knowledge remains a complex challenge. Traditional methodologies such as RAG-based prompting or fine-tuning are frequently employed; however, RAFT offers a more optimal approach for embedding such knowledge. RAFT distinguishes itself by training the model to disregard distractor documents—those that do not contribute to answering the query. By concentrating on pertinent documents and accurately citing the sequences necessary for responding to questions, RAFT substantially enhances the model's reasoning abilities. This chain-of-thought response mechanism refines the model's overall performance. Despite this technique's positive results, producing the training base according to the chain of thought model is difficult and expensive, as another powerful LLM is required.

This work focuses on the MERCOSUR Common Nomenclature (NCM) code system, which is used by all member countries of MERCOSUR. The NCM code system is derived from the Harmonized Commodity Description and Coding System, commonly known as the "Harmonized System" (HS), which is a multipurpose international product nomenclature developed by the World Customs Organization and used by more than 200 countries. The NCM code is widely used in Brazil and is mandatory for all domestic sales operations and international import and export transactions. Due to the complexity of the NCM system, which involves product classification, specifications, and tax ranges, advanced language processing capabilities are required. Ordinary English-Portuguese translation is inadequate for this process, as it requires a sophisticated understanding of Portuguese and precise handling of proper nouns. Preliminary experiments indicate that using only ChatGPT or TeenyTineLLaMA, a Portuguese LLM, is insufficient for effectively processing these tasks.

The simple task of classifying commodity descriptions can be handled with simpler techniques like neural networks by a hierarchical sequence learning \cite{du2021hscodenet}. This study aims not merely to classify product descriptions but to extract the inherent knowledge within the NCM code system. The model can be enriched with specialised knowledge by elucidating the semantic relationships between product descriptions and their respective classifications. Such enhancements hold considerable potential for commerce, compliance, and taxation applications.

This study introduces the ``Simplified Logical Intelligent Model'' - SLIM Retrieval-Augmented Fine-Tuning (SLIM-RAFT) model to effectively address the challenges associated with the NCM code. It considers the constraints of developing a comprehensive training base akin to the original RAFT model. The SLIM-RAFT model can be applied to any element description and multi-classification problems.

A distinguishing feature of the SLIM-RAFT model is its use of a significantly smaller LLM source than traditional models. Specifically, the TeenyTineLLaMA model, comprising only 160 million parameters, was utilised in constructing SLIM-RAFT as the fine-tuned LLM. The results achieved by our model substantially outperform those of ChatGPT 4 in the proposed challenge: ChatGPT 4 scored 4.5/10, whereas SLIM-RAFT achieved an impressive score of 8.67/10. The SLIM-RAFT scheme can be seen in Figure \ref{fig: fig1_slim-raft}.

This work is organized as follows: 

\begin{itemize}
    \item Section 2 presents the works directly related to the ideas proposed in constructing SLIM-RAFT;
    \item Section 3 shows the structure and functioning of the HS and NCM codes;
    \item Section 4 properly demonstrates the construction of the SLIM-RAFT model;
    \item Section 5 presents the results obtained in the comparative evaluations of the models and discusses what was founded;
    \item Section 6 concludes the work and suggests future work related to the presented model.
\end{itemize}

\section{\uppercase{Related Works}}
This section presents related work in two areas: 1) research on the implementation of LLMs with Portuguese as the primary language, and 2) studies on Retrieval-Augmented Generation (RAG) and Retrieval-Augmented Fine-Tuning (RAFT).

\subsection{Portuguese LLM's}

The introduction of LLaMA \cite{touvron2023llama} as an open and efficient foundational language model marked a significant milestone in the evolution of language processing. With models spanning 7 billion to 65 billion parameters, LLaMA underscored the viability of training cutting-edge models exclusively on publicly available datasets. Notably, the LLaMA-13B model surpassed GPT-3 in various benchmarks, demonstrating its remarkable performance despite a comparatively smaller parameter count. This initial success paved the way for subsequent iterations, with LLaMA 2 and 3 \cite{touvron2023llama2,meta2024llama3} further refining the models. These later versions, particularly tailored for dialogue applications, introduced fine-tuned language models optimized for chat interfaces. Consequently, these early advancements laid a robust foundation for future progress in natural language processing.

Although ample data exists for training transformer-based language models in Portuguese, native speakers can readily discern limitations in the text generation and performance of pre-trained models primarily derived from English language data. In recent years, there has been a rising interest in the development and enhancement of large-scale language models tailored to the Portuguese language. This trend has been propelled by pioneering and innovative research efforts, as evidenced by numerous recent contributions to the field.

In European Portuguese (PT-PT), the initiative known as GlórIA \cite{lopes2024gloria} merits particular attention. This project involves a trained decoder language model meticulously constructed from a corpus comprising 35 billion tokens from various sources.

For Brazilian Portuguese (PT-BR), the first relevant LLM encountered was Sabiá \cite{pires2023sabia}. This initiative underscores the development of robust and scalable language models for the Portuguese language. Leveraging advanced machine learning architectures, these models have been instrumental in advancing natural language processing applications in Brazilian Portuguese.

The Cabrita model \cite{larcher2023cabrita} was launched as a low-cost alternative for training LLMs. The authors posited that their methodology could be extended to any transformer-like architecture. To substantiate their hypothesis, they undertook continuous pre-training exclusively on Portuguese text using a 3-billion-parameter model known as OpenLLaMA. This effort culminated in the creation of openCabrita 3B. Remarkably, openCabrita 3B incorporates a novel tokenizer, significantly reducing the number of tokens necessary to represent the text. Subsequently, in a comparable approach, a new study introduced a model predicated on LLaMA 2, designed specifically for handling prompts in Portuguese. This model, named Bode \cite{garcia2024bode}, is available in two versions: one with 7B and 13B parameters. Both models used the LoRa \cite{hu2021lora} fine-tuning method over an open-source LLM. This technique preserves the original parameters intact while introducing a new terminal layer atop the model, which is subsequently trained to achieve the desired fine-tuning outcome.

A noteworthy recent publication entitled ``TeenyTinyLlama: Open Source Tiny Language Models Trained in Brazilian Portuguese'' \cite{correa2024teenytinyllama} offers a valuable perspective on developing compact, open source language models tailored to Brazilian Portuguese. Despite their reduced scale, these models hold significant potential for democratizing access to natural language processing technology, particularly within resource-limited communities.

Collectively, these works signify substantial advancements in implementing language models for the Portuguese language. They underscore the diversity of methodologies and the abundance of resources that bolster research and applications in natural language processing and related fields. These ongoing initiatives are poised to continue influencing the future of language technology for Portuguese speakers globally.

\subsection{Retrieval-Augmented Approach}

Retrieval Augmented Generation - RAG \cite{lewis2020rag} represents a transformative approach to enriching the quality and pertinence of generated content by integrating external insights derived from extensive datasets or repositories of knowledge. By embedding pertinent external knowledge sources into the generation process, RAG is designed to elevate the coherence, factual precision, and overall utility of generated text. This methodology proves advantageous in domains necessitating precise and contextually nuanced content generation, such as question-answering, summarizing, and advanced dialogue systems. By controlling retrieval mechanisms, RAG ensures that the resultant outputs are provided with high accuracy and contextual relevance, thereby advancing the frontiers of natural language processing applications.

In pursuit of enhancing the precision of model responses and mitigating the phenomenon of LLM hallucinations, a novel approach has emerged: Retrieval Augmented Fine-Tuning (RAFT) \cite{zhang2024raft}. This methodology integrates the RAG framework with fine-tuning techniques, empowering models not only to acquire domain-specific knowledge but also to adeptly retrieve and comprehend external contexts crucial for task execution. RAFT introduces the idea of chain-of-thought prompting for building the fine-tuning data set. This prompting technique enables the model's answers to show its reason line with a sequence of arguments, enhancing its explicability.

RAG and RAFT were designed to confront the complexity of tailoring LLMs to specialized domains. Within these realms, the emphasis pivots from general knowledge reasoning to optimizing accuracy \emph{vis-à-vis} a meticulously defined array of domain-specific documents.

\subsection{NCM Data Set}

The ELEVEN data set, ELEctronic inVoicEs in the Portuguese language \cite{di2022eleven}, was meticulously curated to furnish researchers and entrepreneurs with a repository of product descriptions categorized under the Mercosur Common Nomenclature (NCM). This extensive database comprises over a million meticulously labelled records, each scrutinized by taxation experts. These descriptions are short texts, limited to 120 characters, and extracted from authentic electronic invoices documenting purchase and sales transactions.

Labelled datasets are a rare commodity, yet they provide indispensable resources for applications reliant on supervised learning \cite{van2020survey}. The ELEVEN dataset has served as a cornerstone for several noteworthy academic endeavours: 1) the development of a CNN-based system for classifying goods \cite{kieckbusch2021scan}; 2) the creation of data visualization tools aimed at identifying outliers and detecting fraud \cite{marinho2022visual}; and 3) the establishment of a framework utilizing automatic encoders to cluster short-text data extracted from electronic invoices, thereby enhancing anomaly detection within numeric fields \cite{schulte2022elinac}.

\section{\uppercase{HS and NCM Codes }}

In the dynamic realm of international trade, customs brokers, exporters, and importers confront the critical task of accurately classifying goods under the Harmonized System (HS) Code, which underpins the Mercosur Common Nomenclature (NCM) code \cite{valencca2023essays}.

The Harmonized System (HS) is the foundation for customs tariffs and the compilation of international trade statistics in over 200 countries and economies. Beyond these primary functions, the HS is employed for various other purposes, including the monitoring of controlled goods, establishing rules of origin, and facilitating trade negotiations. It is also a crucial component of fundamental customs controls and procedures \cite{wco2024countries}.

The MERCOSUR Common Nomenclature - NCM code system (\emph{Nomenclatura Comun do Mercosul} in Portuguese) is utilized by all member countries of MERCOSUR: Argentina, Brazil, Paraguay, Uruguay, and Venezuela \cite{mercosur2024countries}. In Brazil, including the NCM code is a legal requirement on every electronic invoice issued. Consequently, all commercial transactions, whether domestic or international, must incorporate this code \cite{brazil2024ajuste}.

The Harmonized Commodity Description and Coding System, generally called the ``Harmonized System'' or simply ``HS,'' is a multipurpose international product nomenclature developed by the World Customs Organization \cite{wco2018hs}. The HS structure is shown in Table \ref{tab:tab1_hsstructure}. The NCM, structured hierarchically in the same way as HS, consists of numerical codes assigned to various products, thereby facilitating their identification and categorization in trade agreements and customs processes \cite{mercosul2024ncm}.

\begin{table}[htbp]
\caption{Structure of the HS Codes \cite{wco2018hs}.}
\label{tab:tab1_hsstructure} 
\centering
\begin{tabular}{|c|c|}
  \hline
  2 digit (01-97) & Chapter \\
  \hline
  4 digit (01.01 - 97.06) & Heading \\
  \hline
  6 digit (0101.21 - 9706.00) & Subheading \\
  \hline
\end{tabular}
\end{table}

Table \ref{tab:tab2_hs_apple} shows an HS list cutout, showing the distinction in classification codes for fresh and dried apples. While this differentiation may appear trivial, in an import operation where each type of apple is subject to different tax treatments, an error in code designation can result in significant repercussions. On one hand, the seller faces the risk of tax penalties, while on the other, customs authorities may encounter a loss of revenue.

\begin{table}[htbp]
\caption{Headings 08.08 and 08.13 in the HS \cite{wco2018hs}.}
\label{tab:tab2_hs_apple} 
\centering
\begin{tabular}{|c|l|}
  \hline
  08.08 & Apples, pears and quinces, fresh. \\
  \hline
  0808.10 & - Apples \\
  \hline
  0808.30 & - Pears \\
  \hline
  0808.40 & - Quinces \\
  \hline
  ... &  ... \\
  \hline
  08.13 & \begin{tabular}[c]{@{}l@{}}- Fruit, dried, other than that of \\ 
  headings 08.01 to 08.06; mixtures \\
  of nuts or dried fruits of this Chapter. \end{tabular} \\
  \hline
  0813.10 & - Apricots \\
  \hline
  0813.20 & - Prunes \\
  \hline
  0813.30 & - Apples \\
  \hline
  0808.40 & - Other Fruit \\
  \hline
  0808.50 & \begin{tabular}[c]{@{}l@{}}- Mixtures of nuts or dried fruits of \\
  this Chapter \end{tabular} \\
  \hline
\end{tabular}
\end{table}

Accurate classification of goods is paramount, as it directly impacts taxation, compliance with specific regulations such as sanitary and phytosanitary standards, and eligibility for benefits under international trade agreements. Misclassification can lead to penalties, delays in customs clearance, and substantial financial losses \cite{yadav2023impact}.

The MERCOSUR Common Nomenclature code system adds two digits to the HS structure, making it possible to address new items and their respective subgroups, as seen in Table \ref{tab:tab3_ncmstructure}. This system allows for a more precise and specialized classification of products, enabling more items to be catalogued distinctly by accounting for their specific characteristics.

\begin{table}[htbp]
\caption{Structure of the NCM Codes \cite{mercosul2024ncm}.}
\label{tab:tab3_ncmstructure} 
\centering
\begin{tabular}{|c|c|}
  \hline
  2 digit (01-97) & Chapter \\
  \hline
  4 digit (01.01 - 97.06) & Heading \\
  \hline
  6 digit (0101.21 - 9706.00) & Subheading \\
  \hline
  7 digit (0101.21.1 - 9706.00.9) & Item \\
  \hline
  8 digit (0101.21.10 - 9706.00.90) & Sub-item \\
  \hline
\end{tabular}
\end{table}

The task of classifying product descriptions within the HS or NCM system has been explored in academic literature \cite{du2021hscodenet,kieckbusch2021scan,schulte2022elinac}. Techniques such as neural networks with hierarchical learning and convolutional neural networks (CNNs) have been effectively employed to address this task. Nevertheless, certain challenges associated with this domain remain insufficiently addressed.

One notable challenge is the variability in product descriptions: the same product can be described in multiple ways, and context-dependent synonyms and abbreviations further complicate classification. This complexity makes using LLMs a compelling alternative for interpreting product descriptions. Beyond simple classification, LLMs offer the potential to extract deeper knowledge by identifying relationships between products. Table \ref{tab:tab4_abbreviations} shows some abbreviations that can be easily found.

\begin{table}[htbp]
\caption{Abbreviation Examples.}
\label{tab:tab4_abbreviations} 
\centering
\begin{tabular}{|c|l|}
  \hline
  \textbf{English} &  \\
  \hline
  Coc. 2L & = Coca-Cola 2 Liters \\
  \hline
  P. W. Rice & = Parboiled White Rice \\
  \hline
  \textbf{Portuguese} &  \\
  \hline
  Fr. Desc. & \begin{tabular}[c]{@{}l@{}}= Fralda descartável \\ 
  (\emph{Disposable diaper}) \end{tabular} \\
  \hline
  T. Pap. FDupla & \begin{tabular}[c]{@{}l@{}}= Toalha de Papel Folha Dupla\\ 
  (\emph{Double Ply Paper Towel}) \end{tabular} \\
  \hline
  \textbf{French} &  \\
  \hline
  EDT &  = Eau de Toilette \\
  \hline
  EDP & = Eau de Parfum \\
  \hline
\end{tabular}
\end{table}

Context is crucial in language processing, as it can help resolve ambiguities that often arise when considering abbreviations in isolation. This is where TRANSFORMER-based algorithms shine, as their ability to understand context is key \cite{vaswani2017attention}. For example, the abbreviation ``fr.'' in Portuguese could refer to ``fralda'' (diaper), as shown in Table \ref{tab:tab4_abbreviations}, or it could mean ``fruta'' (fruit). However, when the term ``desc'' (meaning ``descartável'' or disposable) follows, the context effectively resolves the ambiguity between ``fralda'' and ``fruit''.

Therefore, developing LLMs capable of dealing with NCM and HS codes is useful in many fields, especially compliance and tax inspection.

Import and export companies, as well as trading companies in general, are interested in issuing their invoices correctly under current regulations. Errors in product descriptions or their classification according to HS and NCM codes may result in financial penalties or even legal prohibitions of trading.

Customs authorities carefully observe the correct indication of HS and NCM codes, as the tax treatment of those products will be defined through this classification. Misguided indications in these codes can be interpreted as an intention to circumvent the rules to pay less taxes. Of course, fines or other sanctions may be applied to those responsible for the commercial transaction involved.

Therefore, using AI models can improve controlling and correcting the issuance of invoices and related documents \cite{kieckbusch2021scan,marinho2022visual}.

\section{\uppercase{SLIM-RAFT Model}}

The SLIM-RAFT model simplifies RAFT logically and intelligently. Just as RAFT maintains the RAG in its designed form, SLIM-RAFT also maintains the RAG mechanism in its structure, see Figure \ref{fig: fig1_slim-raft}.

The preceding sections have elucidated that constructing the training base in the original RAFT model is an expensive endeavour, frequently necessitating the deployment of another powerful LLM. This substantial cost is predominantly attributable to two features of RAFT: the chain-of-thought reasoning and the inclusion of irrelevant documents. While the concept of learning to disregard irrelevant documents is both valid and logical within the context of RAFT's objectives, it is not a requisite for all applications. This insight prompted the exclusion of this feature in the development of SLIM-RAFT.

SLIM-RAFT retains the chain-of-thought concept in its fine-tuning process, albeit simplified. Instead of using lengthy texts or entire documents as input, the approach employs logical arguments derived from the knowledge base. For instance: 1) element ``a'' belongs to set A; 2) set A is contained within set B; 3) consequently, ``a'' belongs to set B. The next subsection will explain how it was done.

\subsection{FT Database and Prompting}


A theoretical example of a list of arguments for constructing the simplified chain of thought:


\begin{itemize}
    \item Doc. 1: 
    \begin{align}
        a &\in A \label{eq1}
    \end{align}
    
    \item Doc. 2: 
    \begin{align}
        A &\subseteq B \label{eq2}
    \end{align}
    
    \item Doc. 3: Consequently, 
    \begin{align}
        \therefore a &\in B \label{eq3}
    \end{align}
\end{itemize}

This was an application of the training base built for fine-tuning within the idea of the simplified chain-of-thought. See below for a generic example of a prompt:

\begin{verbatim}
    [
      {
        "content":
       [context 1 ]….\n
       [context 2 ]…\n
       [context 3 ]…\n
       [...] … \n
       [context n]… \n
       \n
       Answer the following question 
       using information from the 
       previous context: question",
       "role": "user"
      },
      {
      "content": "response + reasoning 
      based on context",
      "role": "assistant"
    }
  ]
\end{verbatim}


An expert in the NCM code developed a series of question-and-answer sets, complete with their respective arguments. Utilising the open version of ChatGPT 3.5, numerous variations of these questions were generated. Subsequently, a Python script was employed to create thousands of pairs [{question + argument}, {answer + argument}], wherein information derived from the NCM database populated the generic questions.

An example of a generic question used in constructing the NCM base is as follows:

\begin{verbatim}
    [
     {
      "content":
      [product {{product}} has NCM code 
      {{NCM}}], \n
      [which refers to the category 
      {{category}}] \n
      \n
      Answer the following question using 
      information from the previous 
      context: What is the 'category' of
      the 'product' {{product}}?"
      "role": "user"
     },
     {
      "content": "the product {{product}} 
      has the NCM code {{NCM}}, which 
      refers to the category 
      {{category}},
      "role": "assistant"
     }
    ]
\end{verbatim}

A sample record in Portuguese as it was recorded is provided in the Appendix of this work.

Then, for building a data training base in SLIM-RAFT mode, there are three steps:

\begin{enumerate}
    \item A domain expert creates a small question-and-answer set, e.g. \emph{``What is the category of the product 'product'?''};
    \item Construct question-and-answer set variations (an LLM could be used), e.g. \emph{``Could you specify the category to which the product 'product' belongs?''};
    \item Populate de question-and-answer set mask. e.g. \emph{``What is the category of the product 'fresh apple package'?'', ``Could you specify the category to which the product 'fresh apple package' belongs?''}
\end{enumerate}

The total number of records in the data training base will be:

\begin{equation}
    N = q \times v \times n
\end{equation}

Where $q$ is the number of question-and-answer created by the domain expert, $v$ is the number of variations from each question-and-answer unit, and $n$ is the total number of samples from the NCM database.  

As delineated earlier, a notable distinction between SLIM-RAFT and the original RAFT lies in the simplified approach to constructing the fine-tuning base while preserving the chain-of-thought methodology.


\subsection{Source LLM and Fine-tuning}

The LLM source chosen for this work was Teeny TinyLLaMA - TTL, available in two sizes: 460 million and 160 million parameters. Two primary characteristics of TTL guided this selection: its compact size and the training on a corpus in Brazilian Portuguese.

While other source models trained in Brazilian Portuguese exist, as discussed in Section 2, their substantial size can make fine-tuning costly, even when employing optimised techniques such as LoRa \cite{hu2021lora}. In contrast, the compact size of TTL made our fine-tuning process more cost-effective, demonstrating its practicality and potential for wider application.

The Fine-tuning process adjusts all model parameters. The reduced size of TTL facilitated this task. The codes employed were adapted from those provided by the authors of the original TTL paper (available on GitHub \footnote{https://github.com/Nkluge-correa/TeenyTinyLlama}) with minor modifications.


All codes developed for SLIM-RAFT are accessible on SLIM-RAFT's GitHub repository. Both TTL models, 160 million and 460 million parameters, were fine-tuned to create SLIM-RAFT. The 160 million parameter version was used in SLIM-RAFT, while the 460 million parameter version was used for comparative analysis during the final model evaluation.

The SLIM-RAFT GitHub repository \footnote{https://github.com/yurifacanha/ncmrag} is a valuable resource that provides the codes used in this study. This open access not only allows the community to reproduce and assess this experiment but also encourages further collaboration and potential contributions to natural language processing.

\section{\uppercase{Results and Discussion}}

The results were evaluated through a comparative analysis of the responses delivered by the tested models. Three other models were chosen for this comparison: TeenyTinyLLaMA 460m, TeenyTinyLLaMA 460m + NCM fine-tuning, and ChatGPT 4. In the end, four models were tested and evaluated by ChatGPT 4.0:

\begin{itemize}
\item Model 1: TeenyTinyLlama with tiny460M without fine-tuning on the dataset, defined as TTL.
\item Model 2: ChatGPT 4.0, defined as GPT.
\item Model 3: TeenyTinyLlama with fine-tuning on the NCM dataset, defined as NCM-TTL.
\item Model 4: TinyLlama with fine-tuning on the NCM dataset and using SLIM RAFT, defined as SLIM-RAFT.
\end{itemize}

The model's responses were then submitted to ChatGPT-4, which compared the outputs produced with the desired outputs. To ensure impartiality in the evaluation, it is important to note that ChatGPT-4 was not informed of which model each response referred to.

\subsection{Results Presentation}

The evaluation used 100 questions and answers not included in the fine-tuning training base. These 100 questions were presented to various models, and their responses were recorded and compared. ChatGPT-4 assessed the quality of each response, scoring it on a scale from 0 to 10. The final score for each model represents the average of the scores assigned to each response. Table \ref{tab:tab5_results} presents the results of this evaluation. It is clear that Model 4 of SLIM RAFT achieved the best score of 8.63 with a standard deviation of 2.30 across the 100 Q/As.

\begin{table}[htbp]
\caption{Score results between the four models.}
\label{tab:tab5_results} 
\centering
\begin{tabular}{|l|c|c|c|c|}
\hline
\multicolumn{1}{|c|}{\textbf{Model}} & \textbf{Aver.} & \textbf{St. Dev.} & \textbf{Min.} & \textbf{Max.} \\ \hline
TTL                                  & 0.2              & 0.98              & 0             & 5             \\ \hline
NCM-TTL                              & 4.71             & 3.53              & 0             & 10            \\ \hline
GPT                                  & 4.5              & 1.39              & 0             & 5             \\ \hline
SLIM-RAFT                            & 8.63             & 2.30              & 0             & 10            \\ \hline
\end{tabular}
\end{table}

Table \ref{tab:tab6_QA} provides a clear comparison of the responses generated by all models, with the desired response serving as the benchmark for evaluation. This benchmark response sets a standard against which the models' performance can be measured.

\begin{table*}\footnotesize
\centering
\caption{Comparison of Q/A results of four models in both English and Portuguese}
\label{tab:tab6_QA}
\begin{tabular}{|l|l|l|}
\hline
\textbf{Q/A and Models}                                             & \textbf{Sample question - Portuguese}                                                                                                                                                                                                                                                                                                                                                                                                                                                                                                                                                                                                                                                                                                                & \textbf{Sample question - English}                                                                                                                                                                                                                                                                                                                                                                                                                                                                                                                                                                                                                                                \\ \hline
Prompting                                                           & \begin{tabular}[c]{@{}l@{}}Qual a categoria NCM correta para o produto :\\ LIMPADOR VEJA PERF.NAT PERF.SUAVE\\ 1LT? Responda apenas a descrição da\\ categoria . Caso não saiba responda a\\ categoria que mais se aproxima. Não adicione\\ qualquer qualquer outro comentário à sua\\ resposta.\end{tabular}                                                                                                                                                                                                                                                                                                                                                                                                                                        & \begin{tabular}[c]{@{}l@{}}What is the correct NCM category for the\\ product: CLEANER VEJA PERF.NAT PERF.\\ SOFT 1LT? Respond only to the category\\ description. If you don't know, answer the\\ category that comes closest. Do not add any\\ other comments to your answer.\end{tabular}                                                                                                                                                                                                                                                                                                                                                                                     \\ \hline
Expected Answer                                                     & \begin{tabular}[c]{@{}l@{}}Outros agentes orgânicos de superfí­cie\\ (exceto sabões); preparações tensoativas,\\ preparações para lavagem (incluí­das as\\ preparações auxiliares para lavagem) e\\ preparações para limpeza (inclusive multiuso\\ e limpadores), mesmo contendo sabão,\\ exceto os produtos descritos nos CEST\\ 11.001.00, 11.004.00, 11.005.00 e 11.006.00;\\ em embalagem de conteúdo inferior ou igual a\\ 50 litros ou 50 kg.\end{tabular}                                                                                                                                                                                                                                                                                     & \begin{tabular}[c]{@{}l@{}}Other organic surface agents (except soaps);\\ surfactant preparations, washing preparations\\ (including auxiliary washing preparations) and\\ cleaning preparations (including multipurpose\\ preparations and cleaners), whether or not\\ containing soap, except the products\\ described in CEST 11.001.00, 11.004.00,\\ 11.005.00 and 11.006.00; in packaging with\\ a content of less than or equal to 50 liters or\\ 50 kg.\end{tabular}                                                                                                                                                                                                       \\ \hline
\begin{tabular}[c]{@{}l@{}}TeenyTinyLLa-\\ MA\end{tabular}                                                      & \begin{tabular}[c]{@{}l@{}}Não consigo responder com base nas\\ informações fornecidas. No entanto, posso\\ fornecer uma explicação geral de como\\ categorizar produtos NCM, incluindo\\ categorias específicas e suas respectivas\\ categorias.\end{tabular}                                                                                                                                                                                                                                                                                                                                                                                                                                                                                       & \begin{tabular}[c]{@{}l@{}}I am unable to respond based on the\\ information provided. However, I can provide\\ a general explanation of how to categorize\\ NCM products, including specific categories\\ and their respective categories.\end{tabular}                                                                                                                                                                                                                                                                                                                                                                                                                          \\ \hline
ChatGPT 4.0                                                         & Preparações para limpeza                                                                                                                                                                                                                                                                                                                                                                                                                                                                                                                                                                                                                                                                                                                             & Cleaning preparations                                                                                                                                                                                                                                                                                                                                                                                                                                                                                                                                                                                                                                                             \\ \hline
\begin{tabular}[c]{@{}l@{}}TeenyTinyLLa-\\ MA FT PT NCM\end{tabular} & \begin{tabular}[c]{@{}l@{}}A categoria NCM correta é : Agentes\\ orgânicos de superfície (exceto sabões);\\ preparações tensoativas, preparações para\\ lavagem (incluindo as preparações auxiliares\\ para lavagem) e preparações para limpeza,\\ mesmo que contenham sabão, exceto as da\\ posição 34.01.  - Preparações acondicionadas\\ para venda a retalho\end{tabular}                                                                                                                                                                                                                                                                                                                                                                        & \begin{tabular}[c]{@{}l@{}}The correct NCM category is: Organic surface\\ agents (except soaps); surfactant preparations,\\ washing preparations (including auxiliary\\ washing preparations) and cleaning\\ preparations, whether or not containing soap,\\ other than those of heading 34.01. -\\ Preparations packaged for retail sale\end{tabular}                                                                                                                                                                                                                                                                                                                            \\ \hline
SLIM-RAFT                                                           & \begin{tabular}[c]{@{}l@{}}o produto LIMPADOR VEJA PERF.NAT PERF.\\ SUAVE 1LT possui categoria: Sabões, agentes\\ orgânicos de superfície, preparações para\\ lavagem, preparações lubrificantes, ceras\\ artificiais, ceras preparadas, produtos de\\ conservação e limpeza, velas e artigos\\ semelhantes, massas ou pastas para modelar,\\ "ceras para odontologia" e composições para\\ odontologia à base de gesso. - Agentes\\ orgânicos de superfície (exceto sabões);\\ preparações tensoativas, preparações para\\ lavagem (incluindo as preparações auxiliares\\ para lavagem) e preparações para limpeza,\\ mesmo que contenham sabão, exceto as da\\ posição 34.01. - - - Preparações\\ acondicionadas para venda a retalho\end{tabular} & \begin{tabular}[c]{@{}l@{}}The CLEANER VEJA PERF.NAT PERF. SOFT\\ 1LT product has a category: Soaps, organic\\ surface agents, washing preparations,\\ lubricating preparations, artificial waxes,\\ prepared waxes, conservation and cleaning\\ products, candles and similar articles,\\ modeling masses or pastes , "dental waxes"\\ and plaster-based dental compositions. -\\ Organic surface agents (except soaps);\\ surfactant preparations, washing preparations\\ (including auxiliary washing preparations) and\\ cleaning preparations, whether or not\\ containing soap, other than those of heading\\ 34.01. - - - Preparations put up for retail sale\end{tabular} \\ \hline
\end{tabular}
\end{table*}

\subsection{LLM Justification}

It is essential to underscore the potential utility of an LLM specialized in this type of task, as it extends beyond mere classification. Whereas a straightforward input-output classification system is confined to specific subjects and input formats, an LLM system can extract semantic knowledge from the training base and demonstrates flexibility in handling various input formats.

Answering a simple direct question like ``What is the NCM code for the product \emph{fresh apple package}'' is not enough. The question can come in several forms or be embedded in a larger context, for example: ``I don't know the NCM code for \emph{fresh apple package}, can you help me?''.

Another pertinent scenario involves cases of attempted tax evasion. For instance, if the product \emph{fresh apple package} is exempt from taxation while \emph{apple juice package} is subject to tax, it could be misleadingly described in a tax document as \emph{apple j. pack.} but assigned the NCM code for \emph{fresh apples package}, which is tax-exempt. If this discrepancy goes undetected by customs authorities, it could result in a loss of revenue due to the uncollected tax. 

The SLIM-RAFT model can effectively help a system for controlling and inspect documents regarding the NCM Code misuse. But, because of its reduced size, the capability of extracting the right question embedded in a bigger context is limited. Let us consider the following example: 

Portuguese - \emph{Fui na padaria e comprei um suco de laranja, percebi que na nota fiscal aparecia um código chamado NCM, mas estava com a impressão borrada. Qual seria o código impresso?}

English - at the bakery and bought some orange juice, I noticed that a code called NCM appeared on the invoice, but the print was blurry. What would be the printed code?

When presented with this query, our model may not discern the central issue, namely, ``What is the NCM category for orange juice?'' Integrating an additional LLM into the system could mitigate this limitation.

The Few-shot prompting technique \cite{ma2024fewshot,gu2021ppt_fewshot} can enable other large language models (LLMs) to reformulate queries, thereby adapting the context to enhance comprehension by the smaller LLM integrated within the SLIM-RAFT model. This approach is shown above.

Portuguese:
\begin{small}
\begin{verbatim}
Possuo um modelo capaz de responder questões 
diretas, mas não consegue responder quando 
ela não vem nos padrões em que foi treinado. 
O padrão de pergunta correto é "Qual a 
categoria NCM correta para o produto: A?", 
onde A é a descrição do produto.
Sua missão é reformular os prompts para 
que meu modelo entenda:
Exemplo 1:
Pergunta original
na padaria e comprei um suco de laranja, 
percebi que na nota fiscal aparecia um 
código chamado NCM, mas estava com a 
impressão borrada. Qual seria o código 
impresso?
pergunta reforumulada: Qual a categoria 
NCM correta para o produto: suco de laranja?
Exemplo 2:
Pergunta original
Sou um importador de produtos cosméticos e 
vou preencher um documento fiscal para a 
importação do produto "TINT MAXTON 10.0 UN". 
Perdi minha planilha de códigos NCM, então 
pergunto qual NCM usar neste caso?
Pergunta reforumulada: Qual a categoria NCM 
correta para o produto: TINT MAXTON 10.0 UN?

Agora é sua vez, faça o mesmo para as 
perguntas abaixo:

1) Eu sou um comerciante que acabou de 
iniciar as atividades. Vou vender cerveja e 
não sei qual código NCM lançar na nota fiscal. 
Pode me ajudar?
2) Meu filho foi no mercado e comprou uma 
bebida deixando a nota fiscal em cima da 
mesa da sala, onde estava escrito "V. ITAL. 
CORBELLI PRIMITIVO TTO 750 ML". Qual bebida 
meu filho comprou? E qual o código NCM?
3) Sabendo que frutas frescas são isentas 
de impostos e que sucos de frutas 
industrializado são tributados, pergunto se 
o produto Dell Vale Laranja é isento ou 
tributado? E qual o seu código NCM?
\end{verbatim}
\end{small}

ChatGPT output:

\begin{small}
\begin{verbatim}
Aqui estão as perguntas reformuladas para o 
formato adequado ao seu modelo:
1. Pergunta reformulada: Qual a categoria NCM 
correta para o produto: cerveja?
2. Pergunta reformulada: Qual a categoria NCM 
correta para o produto: V. ITAL. CORBELLI 
PRIMITIVO TTO 750 ML?
3. Pergunta reformulada: Qual a categoria NCM 
correta para o produto: Dell Vale Laranja?
\end{verbatim}
\end{small}

English translation:

\begin{small}
\begin{verbatim}
I possess a model capable of answering direct 
questions, but it struggles to respond when 
they do not conform to the patterns in which 
it was trained. The correct question format is 
"What is the correct NCM category for the 
product: A?", where A is the product 
description. Your task is to rephrase the 
prompts so that my model can understand them:
Example 1:
Original question:
I was at the bakery and bought orange juice, 
but I noticed that the invoice had a smudged 
code called NCM. What would be the printed 
code?
Rephrased question: What is the correct NCM 
category for the product: orange juice?
Example 2:
Original question:
I am a cosmetic product importer and need to 
fill out a tax document for the import of the 
product "TINT MAXTON 10.0 UN." I lost my NCM 
codes spreadsheet, so I'd like to know which 
NCM to use in this case.
Rephrased question: What is the correct NCM 
category for the product: TINT MAXTON 10.0 UN?

Now it's your turn; do the same for the 
questions below:

1) I am a merchant who has just started a 
business. I will sell beer and do not know 
Which NCM code should be used on the invoice? 
Can you help me?
2) My son went to the market and bought a 
drink, leaving the invoice on the living room 
table, which read "V. ITAL. CORBELLI PRIMITIVO 
TTO 750 ML". What drink did my son buy? And 
what is the NCM code?
3) Knowing that fresh fruits are exempt from 
taxes and that industrialised fruit juices are 
taxed, I ask if the product Dell Vale Orange 
is exempt or taxed? And what is its NCM code?
\end{verbatim}
\end{small}

ChatGPT output English translation:

\begin{small}
\begin{verbatim}
Here are the questions rephrased to the 
the appropriate format for your model:
1. Rephrased question: What is the correct NCM 
category for the product: beer?
2. Rephrased question: What is the correct NCM 
category for the product: V. ITAL. CORBELLI 
PRIMITIVO TTO 750 ML?
3. Rephrased question: What is the correct NCM 
category for the product: Dell Vale Orange?
\end{verbatim}
\end{small}

Nonetheless, it is important to acknowledge that our model, given its current limited size, may not fully comprehend all contexts. The objective, however, is to expand the model as more resources become available, thereby enhancing its capacity and performance; as the model becomes larger, the need for integration with another model will be suppressed. 

\subsection{Limitations}

The SLIM-RAFT model is a prototype developed to illustrate the original RAFT methodology's simplification and propose its application within the NCM domain. Consequently, this model is a highly specialised tool tailored for its designated task.

It is not recommended for use in ChatBot applications, as TeenyTinyLLaMA (TTL) creators have advised against employing the TTL 160m model for such purposes. The TTL 460m model is recommended for chatbot functionalities.

A simplified chain-of-thought approach is employed when constructing the training base for fine-tuning. However, it's crucial to remember that the involvement of a domain expert is beneficial and necessary for developing the reference lines of reasoning, highlighting the irreplaceable role of human expertise in this process.

\section{\uppercase{Conclusions}}
\label{sec:conclusion}

The SLIM-RAFT model demonstrated significantly superior performance to ChatGPT 4 in interpreting and classifying product descriptions according to the NCM code.

This outcome indicates that a smaller-scale LLM with specific domain knowledge can surpass a more powerful LLM in specialized tasks, provided it is appropriately adjusted and trained while maintaining low execution costs.

The technique for simplifying the construction of the chain-of-thought, as proposed in SLIM-RAFT, not only reduces costs but also proves to be a viable alternative for developing specialized LLMs with high accuracy.

Since NCM coding is used not only for managing, transporting, paying, and taxing various goods in the import and export trade between MERCOSUR countries but also for most tax bills for goods, commodities, and restaurants in the Brazilian market, the practical value of this research is substantial. The findings provide convenience for government departments involved in import and export, taxation, banks, transportation, and manufacturers. Additionally, since over 200 countries use the HS system for import and export trade, the LLM-NCM solution proposed in this article can also facilitate the effective promotion of LLM-HS applications worldwide.

Future research will focus on applying SLIM-RAFT to larger LLMs, such as LLaMA 3, emphasising multilingual tasks. Additionally, comparisons with other fine-tuning techniques, such as LoRa, will be explored.

\section*{\uppercase{Acknowledgements}}

ChatGPT 4 was used in all sections of this work to standardize and improve the writing in British English. This research is partially funded by the Brazilian National Council for Scientific and Technological Development (CNPq).



\section*{\uppercase{Appendix}}

Below, by illustration, a record used in the training base is presented. It is in Portuguese, as it was inserted into the database.

\begin{verbatim}
[
  { "content": " context: 
   [o codigo da categoria do VIN. PORT. 
   QUINTA DE VALE VEADOS RESERVA TT 2014
   750ML é: 22041010],
   [a categoria 22041010 possui a 
   seguinte descrição: Bebidas, líquidos 
   alcoólicos e vinagres. - Vinhos de uvas 
   frescas, incluindo os vinhos 
   enriquecidos com álcool; mostos de uvas, 
   excluindo os da posição 20.09. - - 
   Vinhos espumantes e vinhos espumosos - 
   Tipo champanha (champagne) tem código: 
   22041010],
   [a categoria Bebidas, líquidos 
   alcoólicos e vinagres. - Vinhos de uvas 
   frescas, incluindo os vinhos enriquecidos 
   com álcool; mostos de uvas, excluindo os 
   da posição 20.09. - - Vinhos espumantes 
   e vinhos espumosos - Tipo champanha 
   (champagne) tem posição: Vinhos de uvas 
   frescas, incluindo os vinhos enriquecidos 
   com álcool; mostos de uvas, excluindo os 
   da posição 20.09.] \n 
   responda a seguinte pergunta usando 
   informações do contexto anterior: O 
   produto VIN. PORT. QUINTA DE VALE VEADOS 
   RESERVA TT 2014 750ML está classificado 
   em qual categoria NCM?", "role": "user" 
  }, 
  { "content": "o produto VIN. PORT. QUINTA 
   DE VALE VEADOS RESERVA TT 2014 750ML 
   possui categoria: Bebidas, líquidos 
   alcoólicos e vinagres. - Vinhos de uvas 
   frescas, incluindo os vinhos enriquecidos 
   com álcool; mostos de uvas, excluindo os 
   da posição 20.09. - - Vinhos espumantes 
   e vinhos espumosos - Tipo champanha 
   (champagne)", "role": "assistant" 
  } 
]
\end{verbatim}

\end{document}